\documentclass[sigconf]{acmart}
\renewcommand\footnotetextcopyrightpermission[1]{} % removes footnote with conference information in first column
\AtBeginDocument{%
  \providecommand\BibTeX{{%
    \normalfont B\kern-0.5em{\scshape i\kern-0.25em b}\kern-0.8em\TeX}}}

\acmConference[]{April}{2022}{D-MAE Mocap}
\acmPrice{15.00}
\acmISBN{978-1-4503-XXXX-X/18/06}

\acmSubmissionID{251}

\usepackage{xcolor}
\usepackage{multirow}
\usepackage{algorithm}
\usepackage{algorithmicx}
\usepackage{algpseudocode}
\usepackage{amsmath}
\usepackage{url}
\usepackage{soul}
\usepackage{hyperref}
\usepackage{balance}

\begin{document}

\title[D-MAE Mocap]{A Dual-Masked Auto-Encoder for Robust Motion Capture \\with Spatial-Temporal Skeletal Token Completion}


\author{Junkun Jiang$^1$, Jie Chen$^{1,*}$, Yike Guo$^{1,2}$}
\affiliation{
\institution{$^1$Department of Computer Science, Hong Kong Baptist University, Hong Kong SAR, China}
\institution{$^2$Data Science Institute, Imperial college London, London, UK}
\institution{\{csjkjiang, chenjie, yikeguo\}@comp.hkbu.edu.hk,$~$yg@doc.ic.ac.uk}}

\thanks{$^*$ Corresponding author: Jie Chen}

% \renewcommand{\shortauthors}{Junkun Jiang \& Jie Chen \& Yike Guo}
\renewcommand{\shortauthors}{Jiang et al.}

\begin{abstract}
Multi-person motion capture can be challenging due to ambiguities caused by severe occlusion, fast body movement, and complex interactions. Existing frameworks build on 2D pose estimations and triangulate to 3D coordinates via reasoning the appearance, trajectory, and geometric consistencies among multi-camera observations. However, 2D joint detection is usually incomplete and with wrong identity assignments due to limited observation angle, which leads to noisy 3D triangulation results. To overcome this issue, we propose to explore the short-range autoregressive characteristics of skeletal motion using transformer. First, we propose an adaptive, identity-aware triangulation module to reconstruct 3D joints and identify the missing joints for each identity. To generate complete 3D skeletal motion, we then propose a Dual-Masked Auto-Encoder (D-MAE) which encodes the joint status with both skeletal-structural and temporal position encoding for trajectory completion. D-MAE's flexible masking and encoding mechanism enable arbitrary skeleton definitions to be conveniently deployed under the same framework. In order to demonstrate the proposed model's capability in dealing with severe data loss scenarios, we contribute a high-accuracy and challenging motion capture dataset of multi-person interactions with severe occlusion. Evaluations on both benchmark and our new dataset demonstrate the efficiency of our proposed model, as well as its advantage against the other state-of-the-art methods.
\end{abstract}

\keywords{3D human pose estimation, motion capture, masked auto-encoder, transformer, spatial-temporal encoding}


\begin{teaserfigure}
\centering
\includegraphics[width=1.0\textwidth]{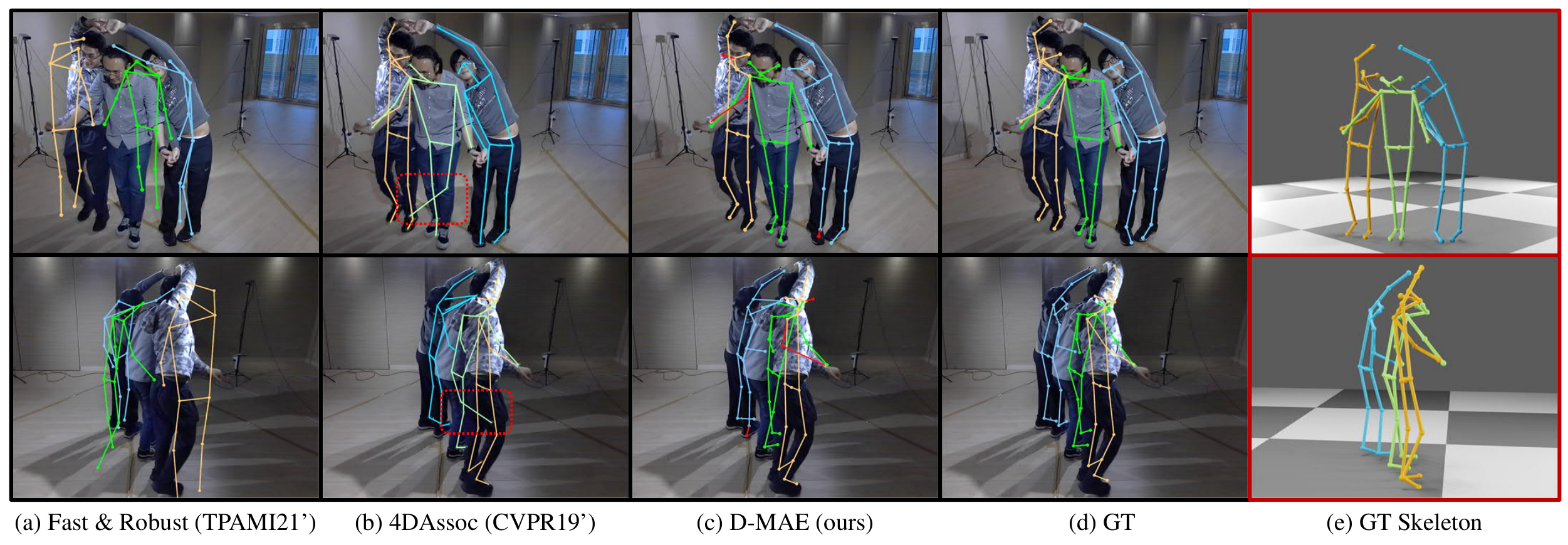}
\caption{Qualitative comparison of the state-of-the-art multi-view multi-person motion capture system and our method on a new motion capture dataset. We show the reprojection of the reconstructed 3D poses with MSCOCO \cite{lin2014microsoft} skeleton definition and also the ground-truth (from right to left, Dong et al. \cite{dong2021fast}, Zhang et al. \cite{zhang20204d}, ours(final), ground-truth). The red joints and limbs indicate the prediction of the proposed D-MAE. We crop and remove visual elements (e.g. bounding boxes) for the clear visualization.}
\label{fig:teaser}
\vspace{1mm}
\end{teaserfigure}

\maketitle

\section{Introduction}

Motion capture (also called Mocap/mo-cap) technology is widely used in film-making, animation, video surveillance, robotics, etc. Commercial motion capture solutions~\cite{optitrack, noitom} adopt multiple sensors to record the target's motion trajectories and coordinates in 3D domain. These techniques are usually inconvenient and costly due to the professional clothing with wearable sensors. Recently, sensor-free methods using multi-view optical-based frameworks have been developed for efficient motion capture at an affordable cost, drawing growing worldwide research attention.

Existing frameworks such as \cite{zhang20204d, lin2021multi, dong2021fast, chu2021part, chen2020cross} build on 2D pose estimations and triangulate to 3D coordinates via reasoning the appearance, trajectory, and geometric consistencies among multi-camera observations referred to as the multi-stage mo-cap system. However, such 2D estimation is usually incomplete and with wrong identity assignments due to limited observation angle, leading to noisy 3D triangulation results and failing in correct motion capture. Thus,
multi-stage mo-cap methods suffer from error-accumulating and encounter a dilemma: it is challenging to retain the correct 2D joints for accurate triangulation and filter the outliers. Strong filtering leads to insufficient data which produces fragmented 3D reconstruction. Specifically, multi-stage mo-cap methods such as \cite{zhang20204d, dong2021fast, lin2021multi} proposed well-designed filters to reduce the gross 2D pose estimation errors, but both of them cannot handle the missing data.
Another line of work tries to fully use the deep learning model instead of the multi-stage pipeline. \cite{tu2020voxelpose, reddy2021tessetrack} gather 2D features across the view to directly associate 3D pose reconstruction. The end-to-end notion provides an elegant pipeline and even allows none-calibrated capturing inputs.
Although these works can achieve high capture accuracy, the computing cost restricts its real-time performance due to its greedy space discretization.

To overcome the above-mentioned issues, our goal is to reason the skeletal structure and motion trajectories to complete the motion sequence:
\textit{\ul{First}}, unlike the previous multi-stage works, we are the first to propose a 3D motion reconstruction and completion framework consisting of an adaptive identity-aware triangulation module called \textit{Adaptive Triangulation} which identically filters false 2D detection via appearance and geometric clues and reconstructs incomplete 3D poses with explicit masking to distinguish the missing joints and,
\textit{\ul{Second}},
we then further propose a Dual-Masked Auto-Encoder (D-MAE) which encodes the joint status with both skeletal-structural and temporal position encoding to generate complete 3D skeletal motion,
% We embed the masked motion sequence dually,
i.e. 1) the 3D coordinate signal encoding, and 2) the joint order and frame timestamp context encoding.
D-MAE's flexible masking and encoding mechanism enable arbitrary skeleton definitions to be conveniently deployed under the same framework.
\textit{\ul{Third}}, in order to demonstrate the proposed model's capability in dealing with severe data loss scenarios, we contribute a high-accuracy and challenging mo-cap dataset of multi-person interactions with severe occlusion.
\textit{\ul{At last}}, comprehensive evaluations on both benchmark and our new dataset demonstrate the efficiency of our proposed model, as well as its advantage against the other state-of-the-art methods.
The code and the dataset will be available online \footnote{\textit{\url{https://github.com/HKBU-VSComputing/2022_MM_DMAE-Mocap}}}.

Our main contributions can be summarized as follows:

\begin{itemize}
\item We propose a Dual-Masked Auto-Encoder (D-MAE) mechanism for the multi-view multi-person motion capturing task. Our proposed framework produces state-of-the-art performances in terms of accuracy and completeness.
\item We propose an adaptive triangulation module for reconstructing identity-aware 3D skeletons with low computational cost, explicitly distinguishing the missing joints for each target identity.
\item Based on the D-MAE's flexible masking and encoding mechanism, we propose a skeleton-fusion strategy for transferring skeletal knowledge between arbitrary skeleton definitions.
\item We propose a large-scale multi-person mo-cap dataset, i.e., BU-Mocap, with complex interactions and heavy occlusions. Accurate manual annotations of 3D skeletal locations are provided. Based on the BU-Mocap, we further validate the advantage of the proposed D-MAE model.
\end{itemize}

The rest of the paper is organized as follows. Sec. \ref{sec:relatedworks} introduces related works, Sec. \ref{sec:method} explains the details of the proposed method, Sec. \ref{sec:evaluation} comprehensively evaluates the proposed model and compares it with the state-of-art methods, as well as ablation studies are carried out, and Sec. \ref{sec:conclusion} concludes the paper.

%%%%%%%%%%%%%%%%%%%%%%%%
\begin{figure*}[t]
\centering
\includegraphics[width=0.96\linewidth]{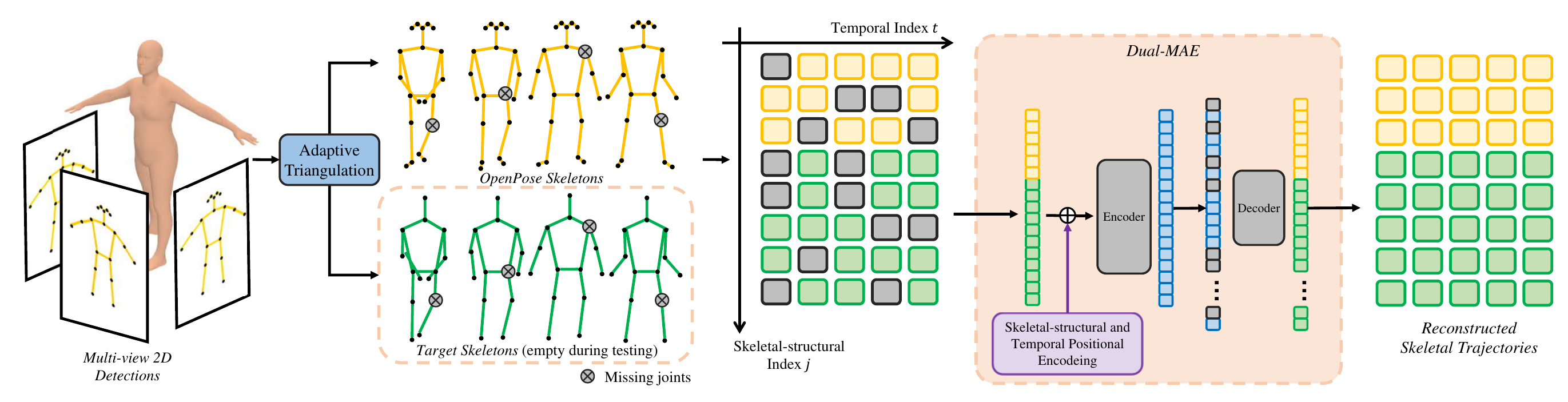}
\caption{System diagram for the proposed D-MAE. We first adopt an off-the-shelf 2D pose detector to estimate 2D poses for each input view. The Adaptive Triangulation module filters 2D noisy joints and reconstructs 3D poses with identification. We encode the input 3D poses via skeletal-structural and temporal position encoding in the proposed D-MAE to generate complete 3D skeletal motion. }
\label{fig:system}
% \vspace{-2mm}
\end{figure*}
%%%%%%%%%%%%%%%%%%%%%%%%

\section{Related Work}\label{sec:relatedworks}

\subsection{Multi-person 3D Motion Capture}
% In this section, we overview literature on 3D multi-view multi-person motion capture.

\textbf{Triangulation-based framework.} Many works have been reported to achieve high-accuracy mo-cap using triangulation-based framework. The main challenge is to establish the cross-view correspondences, i.e. associating 2D poses with the same identity, and it becomes more challenging  when ambiguities caused by severe occlusion are involved. Earlier methods \cite{belagiannis20153d, belagiannis2014multiple, belagiannis20143d} extend the pictorial structure model to implicitly depict the cross-view appearances. It is time-consuming due to its graphical inference and large state space. Recent approaches \cite{zhang20204d, dong2021fast} form a multi-stage trend: it first extracts 2D detection of each view via an off-the-shelf 2D pose estimator. Then it matches 2D poses to get identification across views. Finally, it uses triangulation to reconstruct 3D motion. To utilize temporal information, some tracking algorithms are also added~\cite{belagiannis2014multiple, zhang20204d, dong2021fast}. Another challenge is to filter the 2D failure detection.
With the help of the Part Affinity Fields (PAFs) notion \cite{cao2019openpose}, Zhang et al. \cite{zhang20204d} proposed a bottom-to-top real-time mo-cap system by associating 2D detection under an association energy function. Similar to \cite{cao2019openpose}, the energy function is optimized over the graph edges by parsing the keypoints connectivity over the per-view PAF, cross-view epipolar distance, and cross-frame tracking errors. Dong et al. \cite{dong2021fast} proposed an affinity matrix consisting of 2D appearance features and epipolar geometry to establish the correspondences of the 2D candidates across the view and filter the false detections. They also proposed to model the 3D human skeleton via prior bone length knowledge to reduce reconstruction disturbance over the temporal domain. Slightly different from \cite{zhang20204d, dong2021fast}, Lin et al. \cite{lin2021multi} implement the plane sweep stereo algorithm \cite{collins1996space} via 1D dilated convolutions. They first reconstruct via triangulation only on mid-hip joint for matching and then estimate the 3D skeleton's depth for pair of views regressively. While no clustering methods are utilized in their framework, the reconstructed 3D motion would be duplicated. Additionally, hard distance thresholds for ignoring duplicates will degrade its performance.

However, because of the severe occlusion and noise filtering algorithms, the reconstructed 3D human motions are usually incomplete. The existing methods~\cite{belagiannis20153d, belagiannis2014multiple, belagiannis20143d, zhang20204d, dong2021fast, lin2021multi} cannot handle the incomplete situation leading to the lower performance.

\noindent\textbf{Learning-based framework.} Different from the multi-stage frameworks that primarily rely on triangulation, end-to-end learning models \cite{reddy2021tessetrack, tu2020voxelpose} project the given 2D joint features to the 3D space with volumetric representations and predict 3D motion implicitly. Tu et al. \cite{tu2020voxelpose} proposed 3D-CNN first to detect skeletal 3D bounding boxes and then recover 3D poses intensively in the 3D space. Reddy et al. \cite{reddy2021tessetrack} extend the voxelized framework by a 4D-CNN to extract spatio-temporal representations across the view and over a period of time. It allows to track and reconstruct 3D motion in the wild.

Nevertheless, in spite of the elegant end-to-end manner, directly reconstructing 3D motion via 3D-CNN suffers from large amounts of computations, which is not suitable for low-end devices.

\noindent\textbf{Graph-based Regressing.}
Graph Neural Networks (GNNs) have been proven to be very effective to learn dynamic relations among human motion \cite{zeng2020srnet, ma2021context, ci2020locally, yan2018spatial}. Transformer \cite{vaswani2017attention} can be seen as another form of GNNs attracting researchers to apply its full-attention mechanism to motion reconstruction.
Li et al. \cite{li2021mhformer} adopt Transformer to create multiple 3D pose hypotheses from monocular 2D observation.
Li et al. \cite{li2022exploiting} extend standard Transformer by strided sampling to lift 2D-to-3D poses.
Zheng et al. \cite{zheng20213d} combine 2 vanilla ViTs \cite{dosovitskiy2020image} for 3D reconstruction. Specifically, in every sampling window, the former first encodes each frame's skeletal joints. The latter then treats every encoded feature as a single token to mine the temporal dependencies.
Both of them \cite{li2021mhformer, li2022exploiting, zheng20213d} focus on the monocular 3D pose estimation.
Different from \cite{zheng20213d}'s strategy which uses two separate Transformers, the proposed D-MAE encodes both spatial and temporal information via a dual encoder simultaneously.

\subsection{Motion Trajectory Completion}

Motion completion aims to generate transition clips between given past and future keyframes. The completion unit is a single 3D skeleton. Early pioneer works leverage inverse kinematics introduce space-time constraints to generate physically realistic trajectories \cite{rose1996efficient}. Probabilistic models like Markov \cite{lehrmann2014efficient} and Gaussian process dynamic \cite{wang2007gaussian} models  are widely used to smooth the motion. Harvey et al. \cite{harvey2018recurrent} proposed the Recurrent Transition Networks (RTN) based on the regressive model R-CNN to learn motion representations and then expand it by performing Forward Kinematics to gain a plausible transition results \cite{harvey2020robust}. Recurrent models suffer from computation burdens and range ambiguities. Kaufmann et al. \cite{kaufmann2020convolutional} introduced an end-to-end auto-encoder to fill missing frames. Duan et al. \cite{duan2021single} investigated the BERT \cite{devlin2018bert} and proposed a single-shot network that can predict multiple missing frames within a single forward propagation in real-time. Current in-betweening and in-filling frameworks are designed for short-range motion processing and complete the whole human body.

Nevertheless, existing works cannot complete motion at the joint level. Specifically, different from the above frameworks, we treat every 3D joint (aka keypoint) as a complete unit. The motion sequence of one target is regarded as several joints' collection.

\section{Proposed Method} \label{sec:method}

% just an overview
The pipeline of our proposed framework is shown in Figure \ref{fig:system}. 2D pose estimation is first performed from each camera view using off-the-shelf multi-person pose detector OpenPose \cite{cao2019openpose}, which is trained on the MSCOCO \cite{lin2014microsoft} dataset.
Next, an \textit{Adaptive Triangulation} module is designed to reconstruct identity-aware 3D skeletons from the multi-view 2D poses and identify the missing joints for each identity (Section \ref{triangular}).
Finally, a \textit{Dual-Masked Auto-Encoder} (D-MAE) (Section \ref{MAE}) is employed for skeleton trajectory completion using a transformer-based autoencoder over tokens that encode the joints' 3D locations with both skeletal-structural and temporal position encoding. D-MAE’s flexible masking and encoding mechanism enable arbitrary skeleton annotations to be conveniently deployed under the same framework (Section \ref{ref:fine_tune}). The details of each functional modules will be introduced in the following sub-sections.

\subsection{Adaptive Triangulation}\label{triangular}
To capture 3D motion trajectories for multiple persons, multi-view video cameras are used to capture the same target scene. Based on the multi-view videos, the 2D multi-person pose detector OpenPose \cite{cao2019openpose} is employed to estimate 2D joint locations $\mathcal{D}_{t}^{2d}(v)=\{d_{t, j}^{m}(v) \in \mathbb{R}^2\}$. Here $v\in [1,V] $, $t\in [1,T]$, $j\in[1,J]$, and $m\in[1,M]$ denotes the camera view, video frame, joint index, and person identity, respectively. Meanwhile, a scalar probability value $p_{t, j}^{m}(v)$ is estimated indicating the confidence of the 2D joint estimations.
Experiments show that high confidence 2D joint locations estimated by OpenPose (indicated by $p_{t, j}^{m}(v)$) is highly reliable. However, due to interaction and occlusion among the human targets, some joints could be missing; and some may be assigned to the wrong identity. Inaccuracies also exit during camera parameter calibration. These all cause confusion and error for the subsequent triangulation process leading to (i) inaccurate 3D location estimation, (ii) incorrect cross-view correspondences, and (iii) incomplete skeletal joints reconstruction.
The \textit{Adaptive Triangulation} module is designed to address issues in (i) and (ii); while the D-MAE module introduced in Section \ref{MAE} addresses the inaccuracy issue in (i) and the missing joints problem in (iii).

\setlength{\textfloatsep}{10pt}% Remove \textfloatsep
\begin{algorithm}[t]
\caption{Framework of Adaptive Triangulation.}
\label{alg:Framwork}
\begin{algorithmic}[1]
\Require
The set of 2D skeletons, $\mathcal{D}_{t}^{2d}$;
The set of RGB images from every view, $\mathcal{I}_{t}$;
The set of reconstructed 3D skeletons at the previous frame, $\mathcal{D}_{t-1}^{3d}$;
\Ensure
The set of reconstructed 3D skeletons at the current frame, $\mathcal{D}_{t}^{3d}$;
The map indicating which part is missing, $\mathcal{N}_{t}$;
\State Joint-level Filtering: filter invalid joints;
\State Person-level Filtering: filter 2D skeletons by $\mathcal{D}_{t-1}^{3d}$;
\State Matching: assemble 2D skeletons from each view with the same identity;
\State reconstruct 3D skeletons via triangulation;
\State Labeling: generate the missing map $\mathcal{N}_{t}$ by labeling the missing part of 3D skeletons;
\State store the current 3D skeletons $\mathcal{D}_{t}^{3d}$ for the next frame's Person-level Filtering; \\
\Return $\mathcal{D}_{t}^{3d}$; $\mathcal{N}_{t}$;
\end{algorithmic}
\end{algorithm}

The detail of the Adaptive Triangulation is shown in Algorithm
~\ref{alg:Framwork}, which contains the following steps.

\noindent\textbf{Joint-level Filtering.} We first ensure the reliability of the 2D estimations $\mathcal{D}^{2d}$ by rejecting the ones if (i) the confidence $p_{j}^{m}(v)$ is lower than a given threshold $th_{p}$, and (ii) the number visible cameras is less than $th_{v}$. We found rejecting these noisy inputs help to avoid error accumulation.

\noindent\textbf{Identity Propagation.} When processing multiple frames, the previously reconstructed 3D skeletons, if available, will propagate their identity information to the current frame to avoid re-calculation. We assume the closest $\mathcal{D}_{t-1}^{3d}$ should share the same identity with $\mathcal{D}_{t}^{3d}$. As such, $\mathcal{D}_{t-1}^{3d}$ is projected to each camera view, based on which Euler distances are calculated against the local candidates. If the distance is lower than $th_{d}$, the identity information will be relayed.
The threshold values are set as $th_{p}=0.2$, $th_{v}=3$ and $th_{d}=0.02$ empirically.

\noindent\textbf{Identity Matching.} To associate 2D skeletons from each 2D camera view with the correct identity, we involve both geometric and appearance consistency measures.

The geometry affinity matrix $G_{ij}^{mn}$ is evaluated to indicate the probability that skeleton $m$ and $n$, which are respectively from the camera views $i$ and $j$ share the same identity based on epipolar distances between their respective mid-hip joint locations:

\begin{equation}\label{eq:epipolar}
G_{ij}^{mn} = 1 - \frac{1}{\psi} d_\text{hip}^m(v_i) \oplus d_\text{hip}^n(v_j),
\end{equation}

\begin{equation}
d(v_1) \oplus d(v_2) = \boldsymbol{dist} (\boldsymbol{R}_{v_1}, \boldsymbol{R}_{v_2}),
\end{equation}

where $\boldsymbol{dist}$ stands for the line-to-line distance between $\boldsymbol{R}_{v_1}$ and $\boldsymbol{R}_{v_2}$. $\boldsymbol{R}_{v}$ stands for the ray emitting from the camera center of view $v$ to the point specified by the pixel location $d(v)$. $\psi$ normalizes the affinity scores to the range $[0,1]$.

The appearance affinity matrix $F_{ij}^{mn}$ is similarly defined as the mean squared error between the color features of two image regions:

\vspace{-2mm}
\begin{equation}
F_{ij}^{mn}= 1- \frac{1}{\phi} MSE(\boldsymbol{\Omega}[d_\text{hip}^m(v_i)],\boldsymbol{\Omega}[d_\text{hip}^n(v_j)])
\end{equation}

\begin{comment}
\begin{equation}\label{eq:appearance}
(A_a)_{ij}^{mn} = 1 - \frac{1}{\theta} c^m(v_i) \oplus c^n(v_j),
\end{equation}
\begin{equation}
c(v_1) \oplus c(v_2) = \boldsymbol{dist'} (\boldsymbol{Roi}_{v1}, \boldsymbol{Roi}_{v2}),
\end{equation}
\end{comment}

where $\boldsymbol{\Omega}[d]$ is an operator that calculates the median RGB values of the image region centered around $d$. The median operator improves robustness against light variation. $\phi$ normalizes the distance to the range of $[0, 1]$.

The final matching cost $\mathcal{E}$ is formulated as:

\begin{equation}\label{eq:fuse}
E_{ij}=\sum_{m=0,n=0}^{m=M,n=N}P_{ij}^{mn}*(G_{ij}^{mn}+ F_{ij}^{mn}),
\end{equation}

where $P_{ij}^{mn}$ is the binary ID indicator with value 1 indicate same identity. To find the most possible matching $P$, the energy function $E$ will be maximized using the Hungarian algorithm.

After identity matching and filtering, reconstructed 3D skeletons are incomplete; however, we are fully aware of the exact missing joints for each identity. We set a indicator matrix $N$ and $N_{j}^{m}=1$ indicates the $m$-th candidate's $j$-th 3D joint is missing. $N$ will be used in the D-MAE encoding step.

\subsection{Dual-Masked Auto-Encoder}\label{MAE}

We propose the Dual-Masked Auto-Encoder (D-MAE) for skeletal trajectory completion using a transformer-based autoencoder, which encodes the observed partial skeletal motion trajectories (indicated by the matrix $N$) to a latent space, and reconstructs the original complete trajectories via decoding the latent representations. Inspired by its success in natural language processing and computer vision, we explore its potential in modeling 3D skeletal motion signal which demonstrates unique characteristics and challenges.

\noindent\textbf{Token Indexing and Masking.} If we organize the temporal trajectory of a given joint as a vector, we can treat it as a ``joint patch'' which can be treated similarly to the ``vision patch'' in the Vision Transformer (ViT) \cite{dosovitskiy2020image}. However, the pixels in a vision patch are spatially juxtaposed without explicit autoregressive correlations like those among the temporal joint locations. Therefore, we need a new token definition to reflect the dual directional context along both skeletal-structural (joints dimension) and temporal dimensions of the skeletal motion data.

In the D-MAE, we directly treat a joint's 3D location at a given time as an independent token, and we rely on the transformer-based encoder to explore its skeletal-structural and temporal correlations with the other tokens.
To achieve this, we need two encoders, i.e., a \textit{signal encoder}, and a \textit{context encoder} to transform the relevant information to a higher dimensional space so as to facilitates efficient data fitting by the MLP.
Specifically, the signal encoder $\gamma_s$ encodes each skeletal joint's 3D spatial coordinates $s$; and the context encoder $\gamma_c$ encodes each joint's structural and temporal context $v$. Both are designed as:

\vspace{-2mm}
\begin{equation}\label{eq:signal_encoding}
 \gamma_s(s)= [cos(2\pi\mathbf{B}_s\cdot s), sin(2\pi\mathbf{B}_s\cdot s)]^T,
\end{equation}

\vspace{-2mm}
\begin{equation}\label{eq:contex_encoding}
\gamma_c(v)= [cos(2\pi\mathbf{B}_c\cdot v), sin(2\pi\mathbf{B}_c\cdot v)]^T,
\end{equation}

where each entry in random matrices  $\mathbf{B}_s,~\mathbf{B}_c\in\mathbb{R}^{m\times d}$ are sampled from two normal distributions $\mathcal{N}(0,\sigma_s^2)$ and $\mathcal{N}(0,\sigma_c^2)$, respectively. $\sigma_s$ and $\sigma_c$ are hyperparameters \cite{vaswani2017attention} chosen according to the source signal's distribution.

Let's denote a complete motion sequence for a given target identity as $\mathcal{S}\in\mathbb{R}^{(J\times 3)\times T}$, where $J$ and $T$ denote the total joint number and time frame number, respectively. The input to the D-MAE will be a sequence of \textit{incomplete} skeletal joints' 3D positions along the time axis. A joint is used as input if $N_{j,t}^m=1$, where $j$, $t$ and $m$ are joint, time, and identity indices, respectively. The joint's corresponding encoded token will be calculated as:

\vspace{-2mm}
\begin{equation}\label{eq:full_encoding}
\hat{s}= \gamma_s(s)+\gamma_c(j)+\gamma_c(t).
\end{equation}

Note that, the D-MAE relies on the Adaptive Triangulation module to identify the missing joints information for each identity. To simulate the missing joints situation during model training, we randomly sample a subset of $\mathcal{S}$ as the missing parts and store its index in $N$ under the masking ratio $r$. We use a sliding window with length $T$ to control the model's short-range memory. In practice, we set $T = 15$.

\noindent\textbf{Encoder.} We follow a standard Transformer's \cite{vaswani2017attention} encoder design, which takes the signal- and context-encoded token $\hat{s}$ as input.
These available tokens $\{\hat{s}\}$ will be fed through a series of Transformers blocks into the latent embedding space.

\noindent\textbf{Decoder.}
The input to the D-MAE decoder is the full set of tokens consisting of (i) encoded joint tokens, and (ii) masked missing joint tokens. Each masked missing joint token is a shared, learned vector that indicates the presence of a missing joint to be predicted.
Skeletal-structural and temporal positional encoding as specified in Equation (\ref{eq:contex_encoding}) will be re-applied and added to the \textit{full-set} of tokens, which explicitly encodes the spatial-temporal context to the missing tokens.
Following the asymmetrical design in \cite{he2021masked}, we also design a small decoder which is 50\% narrower and shallower than the encoder.
The output of the decoder will be the \textit{complete} 3D skeletal motion trajectories $\tilde{\mathcal{S}}$.
During training, only the masked joint tokens are supervised by the Mean Squared Error (MSE) loss supervised by the ground truth.

\subsection{Pose Fusion and Fine-tune}\label{ref:fine_tune}

\begin{figure}
\centering
\includegraphics[width=0.85\linewidth]{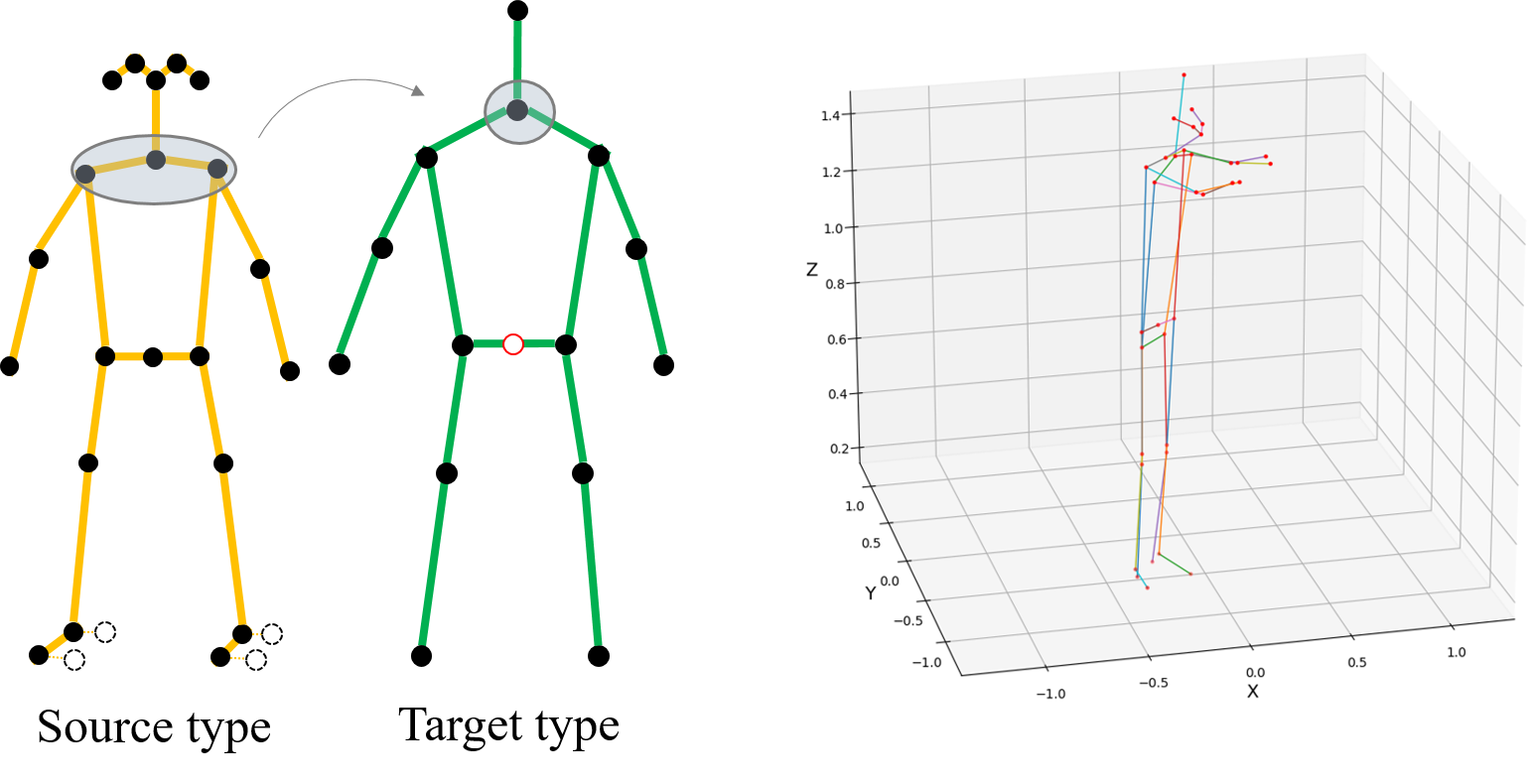}
\caption{Illustration of different skeleton types. Orange one comes from OpenPose \cite{cao2019openpose} definition consists of 25 joints. Green one comes from Shelf \cite{belagiannis20153d} definition consists of 14 joints. The right figure shows 3D visualization. In practice, \textit{big toes} and \textit{small toes} are removed in the orange one and \textit{mid hip} is added in the green one.}
\label{fig:pose_fuse}
% \vspace{-2mm}
\end{figure}

Different skeleton definitions lead to different annotation formats. As shown in the Figure \ref{fig:pose_fuse}, OpenPose \cite{cao2019openpose} defines a human body with 25 joints, while Shelf \cite{belagiannis20153d} defines it with 14 joints. In this section, we try to find the correlation skeleton modelling between two types of definitions.

Existing multi-view multi-person motion capturing systems \cite{dong2021fast, zhang20204d} use interpolation to convert between different skeleton formats. Conversion between the source skeleton type $\mathrm{S}_{src}$ and the target skeleton type $\mathrm{S}_{tar}$ can be formulated as the following:

\begin{equation}\label{eq:skel_conversion}
\mathrm{S}_{src}^{i}=\sum_{j=0}^{j=J}W*\mathrm{S}_{tar}^{j},
\end{equation}

where $\mathrm{S}_{src}\in\mathbb{R}^{3 \times J_{src}}$, $\mathrm{S}_{tar}\in\mathbb{R}^{3 \times J_{tar}}$, and $W\in\mathbb{R}^{J_{src} \times J_{tar}}$ is the interpolation weight. For example, as shown in Figure \ref{fig:pose_fuse}, the target's joint \textit{bottom head} is related to source's joints \textit{right shoulder}, \textit{neck} and \textit{left shoulder}. When one of the source joints \textit{right shoulder}, \textit{neck} or \textit{left shoulder} is missing, the target \textit{bottom head} cannot be interpolated. While people have the ability to inpaint \textit{bottom head} with prior knowledge, e.g. the human kinematic consistency and physical hierarchical body structure. According to this notion, we fine-tune our model by extending the token length of the input motion sequence $\mathcal{S}$: $\mathrm{S}_{src}$ and $\mathrm{S}_{tar}$ are concatenated together as the token unit for every time frame shown in Figure \ref{fig:system}. The output of the D-MAE would be the completed $\mathrm{S}_{src}$ and $\mathrm{S}_{tar}$.

Notice that, our fine-tune task is to use $\mathrm{S}_{src}$ to generate $\mathrm{S}_{tar}$ and simultaneously learn the correlation between the 2 skeleton formats. $\mathrm{S}_{src}$ is prepared by simple triangulation during training. During training, we mask the entire $\mathrm{S}_{tar}$ and partially mask $\mathrm{S}_{src}$ with masking ratio $r_{ft}$, then supervise $\mathrm{S}_{tar}$ only.

\section{Experiments}\label{sec:evaluation}

\subsection{Datasets and Evaluation Metrics}

\noindent\textbf{Datasets.}
The Shelf dataset contains 4 actors' indoor motion with 5 calibrated cameras. While the video sequence consists of 3200 frames, the 3D ground-truth is sparse (e.g., only one annotated actor from frame 378 to frame 499, actor one has about 200 annotated frames, while actor three only has around 30). It is challenging due to its complicated environment, severe occlusion, and the imbalanced and insufficient training data. We follow \cite{belagiannis20143d, belagiannis2014multiple, belagiannis20153d} to prepare the training and testing datasets.

Recently, Joo et al. \cite{joo2015panoptic, Joo_2017_TPAMI} established a massive indoor capturing dataset called CMU Panoptic with variant actors performing solo and social activities. The multi-view cluster system provides high-definition RGB frames and depth data. However, without manual correction, the reconstructed ground-truths have several incorrect annotations, and there is a small number of close interactions and challenging motion.
Zhang et al. \cite{zhang20204d} contributed a dataset captured by the commercial motion capture optical-based system OptiTrack \cite{optitrack}. With the help of commercial cameras, passive markers and high-performance post-processing algorithms, the ground-truth in Zhang et al. \cite{zhang20204d}'s dataset is of high accuracy.  Due to that all the actors wear marker-attached black tight suits for less appearance, it is unfair to use for evaluating appearance-based approaches.

\begin{figure}
\centering
\includegraphics[width=0.98\linewidth]{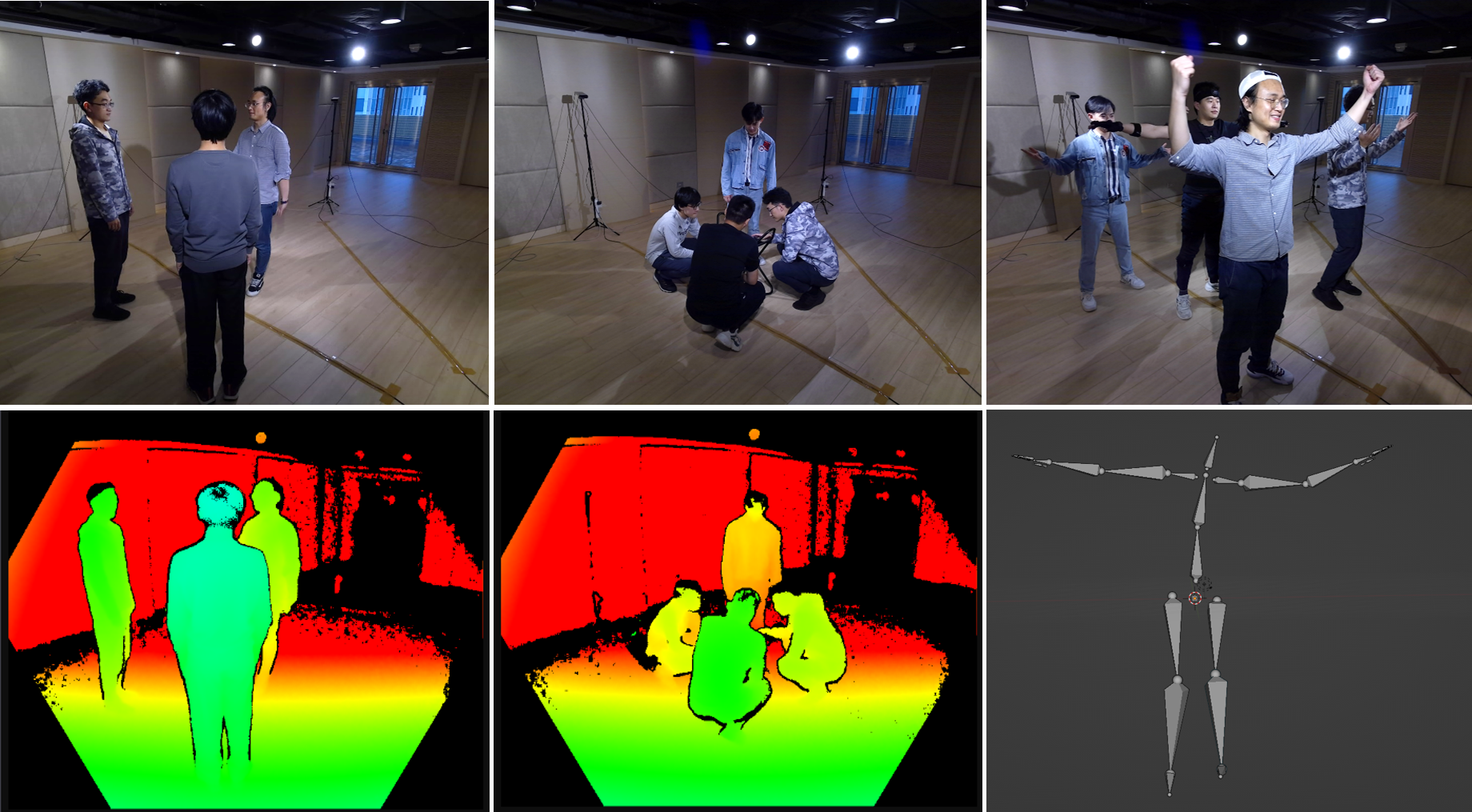}
\caption{Illustration of our multi-modal motion capturing dataset. (Top) 2D observation of sample scene. (Bottom) depth map captured from Azure Kinect \cite{azurekinect} and motion trajectory captured from Noitom \cite{noitom}.}
\label{fig:dataset}
% \vspace{-2mm}
\end{figure}

To this end, we contribute the BU-Mocap, a new dataset of multi-person interactions with severe occlusions. We develop a multi-view multi-person annotation system to manually correct the wrong 2D detection as well as 3D reconstruction. As shown in Figure \ref{fig:dataset}, we use 5 synchronized commercial cameras Azure Kinect \cite{azurekinect} and IMU sensor Noitom PN3 \cite{noitom} to record live motion with point cloud data and joint trajectories. Our BU-Mocap dataset \footnote{We will release the recording point cloud and multi-modal motion data in the future, and we will extend it for variant motion capturing. Note that in this paper, we only utilize the vision frames for training and testing.} contains 5 motion clips with 1000 frame-length capturing for each and 4-5 actors' interactions under 30 FPS.
In brief, we first apply the proposed adaptive triangulation module to reconstruct 3D motion coarsely and then project the 3D poses backward to 2D frames to manually correct the keypoints for each actor.

\noindent\textbf{Metrics.}
Based on the evaluation protocol of previous works \cite{zhang20204d, lin2021multi, reddy2021tessetrack, lin2021multi}, we use the Percentage of Correctly estimated Parts (PCP), Percentage of Correctly estimated Keypoints (PCK), and Mean Per Joint Position Error (MPJPE) for the evaluation. PCP indicates the correctness of the predicted limb via comparing the distance of two predicted joints and the length of the ground-truth limb (consist of two joints). PCK shows the accuracy in joint level. The prediction is true if the distance between the predicted position and the ground-truth in a certain threshold. In this case, we set the threshold at 0.2 meters and use PCK@0.2 to calculate the Precision-Recall metric. MPJPE shares a similar notion to PCK, which can evaluate the accuracy of 3D reconstruction via Euler distance. The unit of MPJPE is millimeter.

\noindent\textbf{Implementation Details}
At first, all joints are subtracted by the skeleton center \textit{mid hip} in XYZ coordinate system.
We refer to this subtract procession as \textit{Normalization}. To handle the imbalance and insufficient annotations and avoid overfitting, we randomly rotate the training data along z-axis after \textit{Normalization}. Using both \textit{Normalization} and random rotation improves the proposed D-MAE's performance.

We implement our D-MAE based on a standard Transformer \cite{vaswani2017attention}.
We adopt Fourier positional encoding \cite{tancik2020fourier} for XYZ signal encoding and context encoding (a.k.a., skeletal-structural position encoding and temporal position encoding).
In the training process, all the transformer modules use a dropout rate \cite{srivastava2014dropout} of $0.1$. We adopt AdamW \cite{paszke2019pytorch} optimizer with a cosine decay learning rate began with $10^{-5}$ to train the model for 1000 epochs with a batchsize of 32. In the fine-tune process, we train the model in the same manner. The number of fine-tune epoch is set to 500. For more implementation details, please refer to supplementary.

\setlength{\tabcolsep}{4pt}
\begin{table}
\begin{center}
\caption{Comparison of PCP(\%) metric on the Shelf dataset with existing state-of-the-art multi-view multi-person motion capture methods. We separate Actor-2 due to its obvious inaccurate annotations. `AVG' means the average of PCP. `$\dagger$' means Actor-2 is excluded.}
\label{table:shelf_PCP}
\begin{tabular}{rccccc}
\hline\noalign{\smallskip}
Method & Actor-1 & Actor-2 & Actor-3 & AVG & AVG$\dagger$ \\
\hline
\noalign{\smallskip}
Belagiannis et al.~\cite{belagiannis20143d}       & 66.1          & 65.0          & 83.2          & 71.4          & 74.7          \\
Belagiannis et al.~\cite{belagiannis2014multiple} & 75.0          & 67.0          & 86.0          & 76.0          & 80.5          \\
Belagiannis et al.~\cite{belagiannis20153d}       & 75.3          & 69.7          & 87.6          & 77.5          & 81.5          \\
Dong et al.~\cite{dong2021fast}                   & 98.8          & 94.1          & 97.8          & 96.9          & 98.3          \\
Zhang et al.~\cite{zhang20204d}                   & 99.0          & 96.2          & 97.6          & 97.6          & 98.3          \\
Reddy et al.~\cite{reddy2021tessetrack}           & 99.1          & 96.3          & 98.3          & \textbf{98.2} & 98.7          \\
Lin et al.~\cite{lin2021multi}                    & 99.3          & \textbf{96.5} & 98.0          & 97.9          & 98.7          \\
Ours w/o D-MAE                                    & 97.1          & 88.6          & 98.2          & 94.7          & 97.7          \\
Ours w/o FT                                       & 97.3          & 89.7          & 98.2          & 95.0          & 97.8          \\
\hline
\noalign{\smallskip}
Ours(final)                                       & \textbf{99.7} & 94.1          & \textbf{98.4} & 97.4          & \textbf{99.1} \\
\hline
\end{tabular}
\end{center}
% \vspace{-2mm}
\end{table}
\setlength{\tabcolsep}{1.4pt}

%%%%%%%%%%%%%%%%%%%%%%%%
\begin{figure*}[t]
\centering
\includegraphics[width=1.\linewidth]{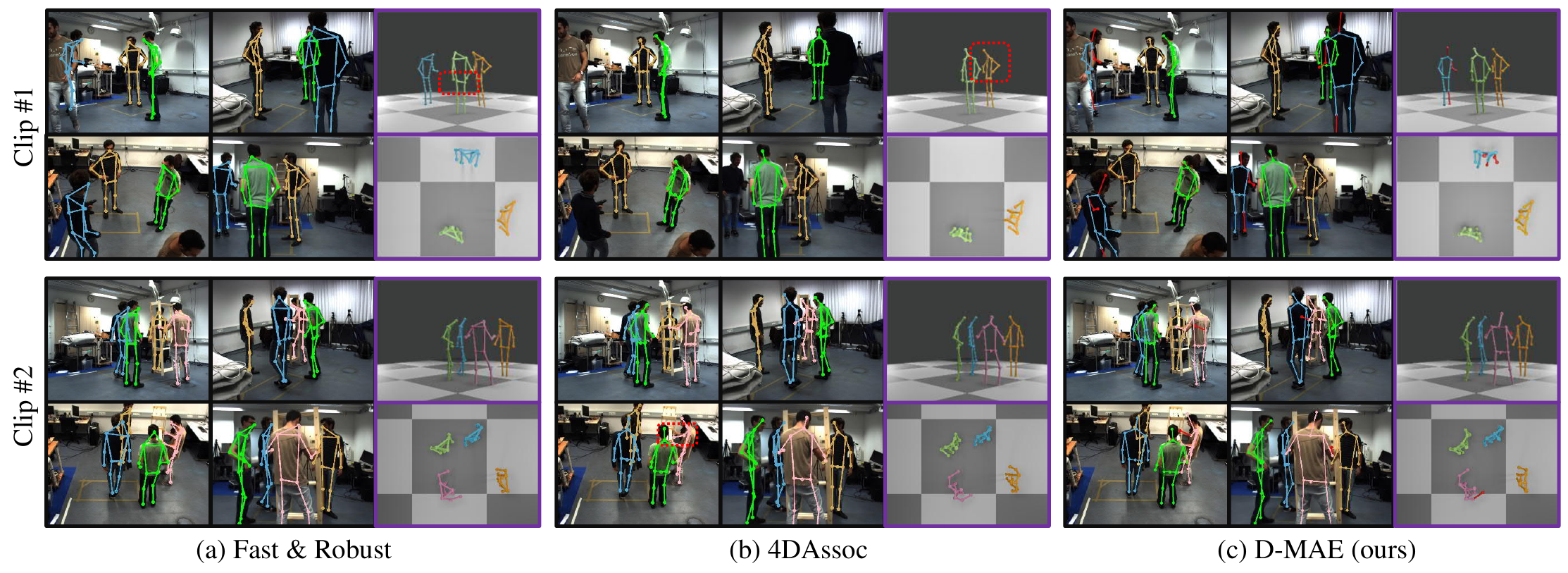}
\setlength{\abovecaptionskip}{-3mm}
\setlength{\belowcaptionskip}{-1.5mm}
\caption{Qualitative evaluation on Shelf dataset. We show reprojection of 3D pose on four views. The right column shows a side view rendering and a top view rendering for 3D visualization. Joints and limbs in red color demonstrate the predictions from the proposed MAE model. Thus, after filtering 2D detections in low confidence, triangulation reconstructs incomplete 3D skeletons. Our MAE model generates the missing parts to complete the 3D pose.}
\label{fig:shelf_qualitative}
\end{figure*}

\begin{figure*}[t]
\centering
\includegraphics[width=0.95\linewidth]{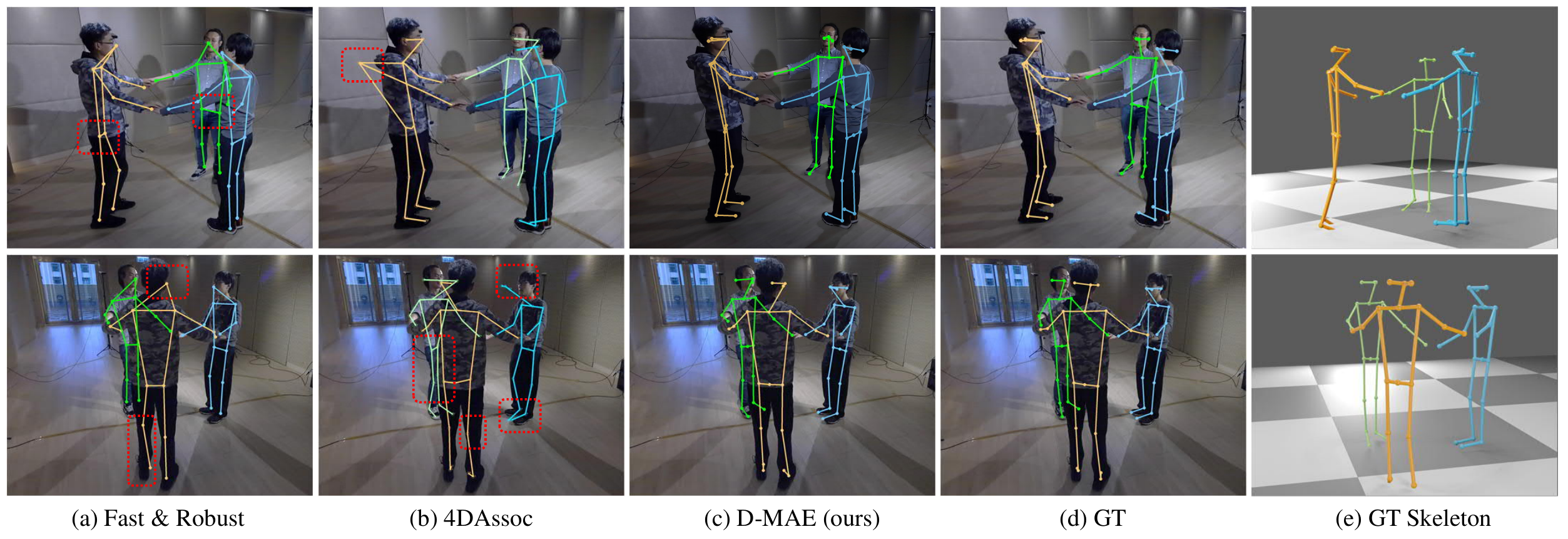}
\setlength{\abovecaptionskip}{1mm}
\caption{Qualitative evaluation on our proposed X-Mocap dataset. We show reprojection of 3D pose on four views. The red dotted boxes are highlights for attention.}
\label{fig:xmocap_qualitative}
\end{figure*}
%%%%%%%%%%%%%%%%%%%%%%%%

\setlength{\tabcolsep}{4pt}
\begin{table*}
\begin{center}
\setlength{\abovecaptionskip}{1mm}
\setlength{\belowcaptionskip}{-2mm}
\caption{Comparison on insufficient views on Shelf dataset. As the number of cameras decreases, performance of different methods decrease due to insufficiency.}
\label{table:shelf_view_PCP}
\begin{tabular}{r|ccc|ccc|ccc|ccc}
\hline
Method \textbackslash PCP(\%)         & \multicolumn{3}{c|}{Actor-1}                                                            & \multicolumn{3}{c|}{Actor-2}                                                            & \multicolumn{3}{c|}{Actor-3}                                                            & \multicolumn{3}{c}{Average}                                                             \\ \hline
Cameras number                        & \multicolumn{1}{c|}{3}             & \multicolumn{1}{c|}{4}             & 5             & \multicolumn{1}{c|}{3}             & \multicolumn{1}{c|}{4}             & 5             & \multicolumn{1}{c|}{3}             & \multicolumn{1}{c|}{4}             & 5             & \multicolumn{1}{c|}{3}             & \multicolumn{1}{c|}{4}             & 5             \\ \hline
Zhang et al.~\cite{zhang20204d}       & \multicolumn{1}{c|}{92.8}          & \multicolumn{1}{c|}{97.3}          & 98.8          & \multicolumn{1}{c|}{68.1}          & \multicolumn{1}{c|}{83.2}          & 94.1          & \multicolumn{1}{c|}{79.9}          & \multicolumn{1}{c|}{95.2}          & 97.8          & \multicolumn{1}{c|}{80.3}          & \multicolumn{1}{c|}{91.9}          & 96.9          \\ \hline
Lin et al.~\cite{lin2021multi}     & \multicolumn{1}{c|}{\textbf{98.7}} & \multicolumn{1}{c|}{98.4}          & 99.3          & \multicolumn{1}{c|}{88.4}          & \multicolumn{1}{c|}{\textbf{96.0}} & \textbf{96.5} & \multicolumn{1}{c|}{\textbf{98.1}} & \multicolumn{1}{c|}{98.0}          & 98.0          & \multicolumn{1}{c|}{95.1}          & \multicolumn{1}{c|}{\textbf{97.5}} & \textbf{97.9} \\ \hline
Ours(final)                           & \multicolumn{1}{c|}{96.8}          & \multicolumn{1}{c|}{\textbf{99.2}} & \textbf{99.7} & \multicolumn{1}{c|}{\textbf{93.8}} & \multicolumn{1}{c|}{92.2}          & 94.1          & \multicolumn{1}{c|}{97.3}          & \multicolumn{1}{c|}{\textbf{98.3}} & \textbf{98.4} & \multicolumn{1}{c|}{\textbf{96.0}} & \multicolumn{1}{c|}{96.6}          & 97.4          \\ \hline
\end{tabular}
\end{center}
\vspace{-3mm}
\end{table*}
\setlength{\tabcolsep}{1.4pt}

\vspace{-3mm}
\subsection{Quantitative Evaluation}

We first compare our method with the state-of-the-art multi-view multi-person 3D approaches \cite{belagiannis20143d, belagiannis2014multiple, belagiannis20153d, dong2021fast, zhang20204d, reddy2021tessetrack, lin2021multi} on the Shelf \cite{belagiannis20153d} dataset. The qualitative evaluation with PCP metric is presented in Table \ref{table:shelf_PCP}. For Actor-1 and Actor-3, our method can achieve the state-of-the-art performance, even better than the fully learning-based approach \cite{reddy2021tessetrack}. We notice that obvious degradation  can be seen in Actor-2.
In fact, after carefully reviewing the whole clips of Shelf \cite{belagiannis20153d} dataset, we found that some of the annotations of the Actor-2 are inaccurate.
This is also one of the reasons for us to establish a new dataset for the multi-view multi-person interactive task.

We also evaluate two triangulation-based 3D mo-cap systems \cite{zhang20204d, dong2021fast}, one learning-based method \cite{lin2021multi} and ours on the BU-Mocap dataset. We follow \cite{zhang20204d} to use PCK@0.2 to calculate the Precision (the ratio of correct joints in all reconstructed joints) and the Recall (the ratio of correct joints in all ground-truth joints). We unify those 3D poses into the MSCOCO \cite{lin2014microsoft} definition. As shown in Table \ref{table:our_dataset}, our method outperforms \cite{zhang20204d, dong2021fast, lin2021multi}. To demonstrate the D-MAE's generalization, we first evaluate
it which is trained on Shelf (row 4) and fine-tune it on BU-Mocap (row 6). Zhang et al. \cite{zhang20204d} achieve very close performance to ours on Precision and Recall, while ours outperforms on PCP indicating our 3D reconstruction is closer to reality. Dong et al. \cite{dong2021fast} fall to the bad reconstruction due to the failure of 2D pose matching during their Re-id process. We notice [17]'s heavy degradation (row 3). For the fair comparison, we fine-tune (train on Shelf then on BU-Mocap) both D-MAE and \cite{lin2021multi} (row 5 and 6). Fine-tuned \cite{lin2021multi} falls on the BU-Mocap. We believe that this is caused by the hard distance threshold for estimating the depth volume. Each scene has its unique distance threshold indicating the capturing volume.
If this parameter is set inappropriately, the depth estimation will be incorrect leading to bad reconstruction.

\subsection{Qualitative Evaluation}
\begin{figure}[t]
\centering
\includegraphics[width=0.98\linewidth]{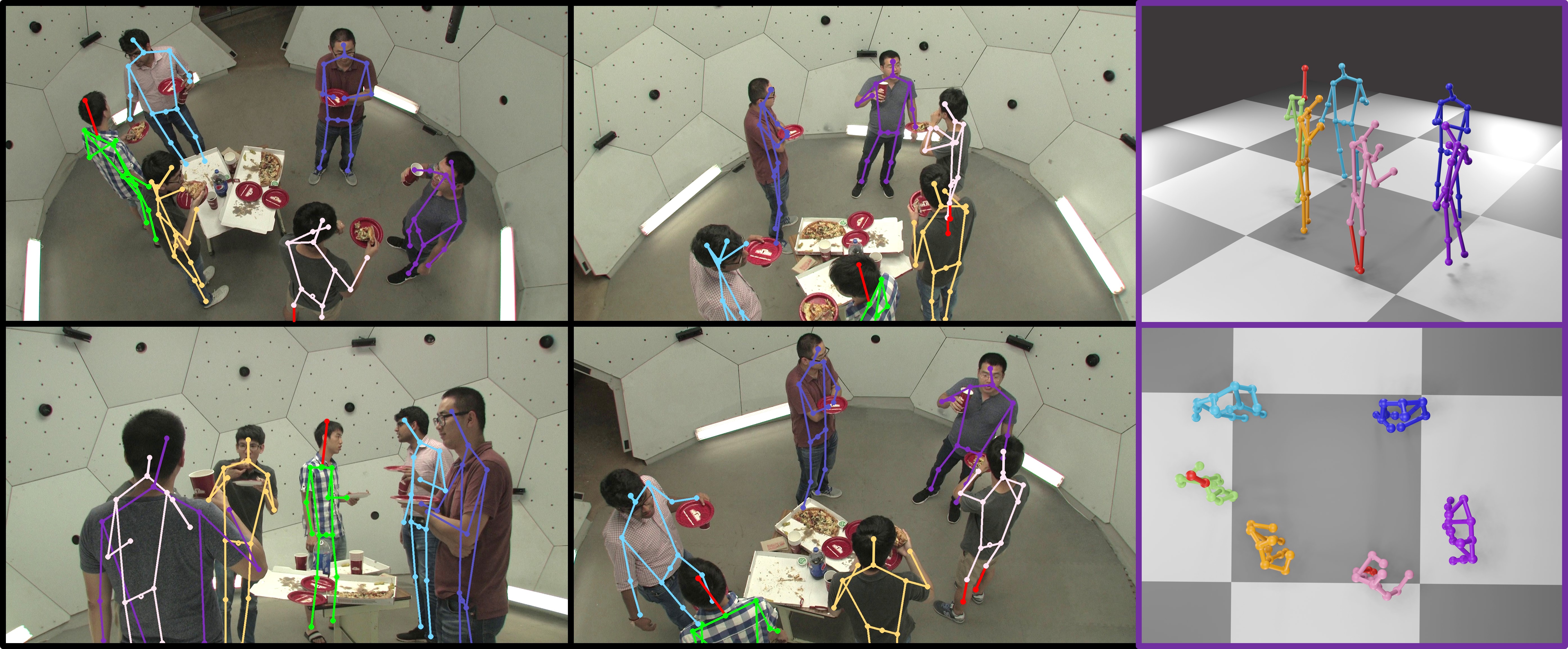}
\caption{Qualitative evaluation on the CMU Panoptic dataset. We show reprojection of 3D pose on four views. The red joints and limbs indicate the prediction of D-MAE.}
\label{fig:panoptic_qualitative}
% \vspace{-3mm}
\end{figure}

We show examples of qualitative results in Figure \ref{fig:teaser}, \ref{fig:shelf_qualitative}, \ref{fig:xmocap_qualitative}, and \ref{fig:panoptic_qualitative}. The false 2D detection from the bottom-to-top pose estimator \cite{cao2019openpose} is filtered by the Adaptive Triangulation module leading incomplete 3D reconstruction. The proposed D-MAE completes the missing 3D joints. The missing 3D joints are marked with red color. Specifically, as shown in Figure \ref{fig:shelf_qualitative}, \cite{dong2021fast, zhang20204d} reconstruct twisted 3D poses on the Shelf dataset. As shown in Figure \ref{fig:teaser} and Figure \ref{fig:xmocap_qualitative}, both of the two frameworks cannot achieve robust performance. \cite{dong2021fast} needs a warm-up, i.e. identity matching would become more accurate during the procession. \cite{zhang20204d} suffers from the temporal tracking error i.e. when the actor rotates quickly, the reconstructed poses would twisted. We also provide the qualitative evaluation on CMU Panoptic dataset, as shown in Figure \ref{fig:panoptic_qualitative}. From Figure \ref{fig:panoptic_qualitative} we can see that, our proposed system even can reconstruct the invisible joints due to the D-MAE completion.

\setlength{\tabcolsep}{4pt}
\begin{table}
\begin{center}
\setlength{\abovecaptionskip}{2mm}
\caption{Comparison with \cite{dong2021fast, zhang20204d, lin2021multi} and ours on BU-Mocap. We choose one scene with 3 person performance. One of the visual example frames is shown as Figure \ref{fig:teaser}. `*' means we fine-tune the model on this scene.}
\label{table:our_dataset}
\begin{tabular}{rccc}
\hline\noalign{\smallskip}
BU-Mocap                         & Precision (\%) & Recall (\%)   & PCP (\%)      \\ \hline
Dong et al. \cite{dong2021fast}  & 84.3           & 75.1          & 78.7          \\
Zhang et al. \cite{zhang20204d}  & 94.7           & \textbf{91.7} & 85.9          \\
Lin et al. \cite{lin2021multi}   & 45.6           & 45.6          & 40.5          \\
Ours(final)                      & \textbf{94.8}  & 90.0          & \textbf{95.2} \\ \hline
Lin et al. \cite{lin2021multi} * & 70.8           & 70.8          & 68.6          \\
Ours(final) *                    & \textbf{97.8}  & \textbf{92.4} & \textbf{97.7} \\ \hline
\end{tabular}
\end{center}
\vspace{-2mm}
\end{table}
\setlength{\tabcolsep}{1.4pt}

\subsection{Ablation Study}

\noindent\textbf{D-MAE completion.} We first evaluate the motion completion capability of the proposed D-MAE model. The result is labeled as ``w/o D-MAE'' in Table \ref{table:shelf_PCP}. Under severe occlusion, only relying on triangulation leads to a substantial down-graded performance, especially the adaptive triangulation framework filters most of the low confidence 2D detection (Actor-2 is covered by a messy environment in a long time period). The proposed D-MAE model completes the missing parts and allows the whole system to gain robust 3D motion capture results.

\setlength{\tabcolsep}{4pt}
\begin{table}
\begin{center}
\caption{Ablation study for Pose Fusion Fine-tune strategy. We report each limb's average PCP(\%) score of Actor-1 to Actor-3 on the Shelf dataset. }
\label{table:ft}
\begin{tabular}{rcc}
\hline\noalign{\smallskip}
Limb name       & w/o FT & final \\
\hline
\noalign{\smallskip}
Left Upper Arm  & 98.9  & \textbf{99.8}   \\
Right Upper Arm & 97.7  & \textbf{98.0}   \\
Left Lower Arm  & 84.6  & \textbf{86.4}   \\
Right Lower Arm & 86.2  & \textbf{94.5}   \\
Left Upper Leg  & 100.0 & 100.0           \\
Right Upper Leg & 100.0 & 100.0           \\
Left Lower Leg  & 99.2  & \textbf{100.0}  \\
Right Lower Leg & 100.0 & 100.0           \\
Head            & 84.3  & \textbf{95.5}   \\
Torso           & 100.0 & 100.0           \\
\hline
\noalign{\smallskip}
Average         & 95.1  & \textbf{97.4}   \\
\hline
\end{tabular}
\end{center}
\vspace{-2mm}
\end{table}
\setlength{\tabcolsep}{1.4pt}

\setlength{\tabcolsep}{4pt}
\begin{table}
\begin{center}
\caption{Effect of the dual directional encoding.
% ``w/o'' means ``without''.
``J-encoding'' and ``T-encoding'' mean the skeletal-structural position encoding and temporal encoding respectively.}
\label{table:dual_encoding}
\begin{tabular}{rcc}
\hline\noalign{\smallskip}
Method       & Recall (\%) $\uparrow$ & MPJPE (mm) $\downarrow$ \\
\hline
\noalign{\smallskip}
w/o J-encoding    & 91.8          & 71.3          \\
w/o T-encoding    & 96.6          & 45.7          \\
w/o J\&T-encoding & 87.2          & 82.5          \\
\hline
\noalign{\smallskip}
baseline           & \textbf{98.7} & \textbf{41.0} \\
\hline
\end{tabular}
\end{center}
% \vspace{-2.5mm}
\end{table}
\setlength{\tabcolsep}{1.4pt}

\noindent\textbf{Pose Fusion fine-tune strategy.} We compare Pose Fusion Fine-tune strategy on the Shelf dataset. Labels ``w/o FT'' and ``final'' stand for D-MAE, which disables and enables fine-tune strategy, respectively. As shown in Table \ref{table:ft}, after fine-tuning, our final model performs much better than ``w/o FT''. Note that the fine-tune strategy has increased the PCP score on limb \textit{Head} about 11\%. The obvious improvement demonstrates that the pose-fuse notion can let the model learn the skeletal conversion and also shows the advantage of the D-MAE's flexible masking and encoding mechanism.

\noindent\textbf{The dual directional encoding.} To demonstrate the effectiveness of the dual directional encoding,
we remove one or all of them to train a D-MAE model without Pose Fusion (a.k.a., ``w/o FT'') on the Shelf training dataset and make the comparison on the Shelf testing dataset via Recall and MPJPE. We take ``w/o FT'' as baseline. We adopt the same training configuration of ``w/o FT'' introduced in Section \ref{ref:fine_tune}. During the testing, the mask ratio on joint level and frame level are both set to $0.5$. As shown in Table \ref{table:dual_encoding}, baseline outperforms other variants. This indicates our model learns both the skeletal structure and the motion trajectory via the dual directional encoding mechanism.

\noindent\textbf{Number of camera views.} We report the performance impact of the number of observations in Table \ref{table:shelf_view_PCP}. We compare the proposed framework with \cite{zhang20204d, lin2021multi}. While all the methods are based on triangulation, the result of \cite{zhang20204d} degrades significantly, especially for Actor-2 (96.5\% in 5 views, 88.4\% in 3 views). Severe loss of observations results \cite{zhang20204d} to treat 2D detection as noise. \cite{lin2021multi} reconstructs 3D motion via plane sweep stereo. Duplicates allow \cite{lin2021multi} to find a closest 3D reconstruction from the ground-truth, while ours performance is much more robust (average degrading ratio: ours is 1.4\%, \cite{lin2021multi} is only 2.8\%).

\section{Conclusions} \label{sec:conclusion}

In this paper, we propose a novel multi-view multi-person 3D motion capture approach. Different from the multi-stage triangulation-based methods, our key idea is to use autoregressive Transformer to perform joint-level motion completion. We first propose the Adaptive Triangulation module to associate and reconstruct 3D pose with the masking map. Then we propose the D-MAE to generate motion with both skeletal-structural and temporal position encoding. To overcome the annotation gap, we leverage Pose Fusion fine-tune strategy and also make our framework adaptable to arbitrary skeleton definitions. Furthermore, we contribute a new dataset captured for multi-view multi-person interactive scene.

\vspace{-0.5mm}
\section*{Acknowledgments}
The research was supported by the Theme-based Research Scheme, Research Grants Council of Hong Kong (T45-205/21-N).

\bibliographystyle{ACM-Reference-Format}
\balance
\bibliography{2022.04_ACM_Mocap}

\end{document}